\pdfoutput=1
\documentclass[11pt]{article}

\usepackage[a4paper,margin=20mm]{geometry}
\usepackage[T1]{fontenc}
\usepackage{amsmath,amssymb,amsfonts}
\usepackage{mathtools}
\usepackage{array}
\usepackage{makecell}
\usepackage{graphicx}
\usepackage{booktabs}
\usepackage{cite}
\usepackage[hidelinks]{hyperref}
\usepackage{microtype}

\newcommand{\Fp}{\mathbb{F}_p}

\newcommand{\Hcube}{\{0,1\}^m}
\newcommand{\MLE}[1]{\widetilde{#1}}

\begin{document}

\title{Agree on the Model, Verify the Inference:\\
GKR Protocols for HND-Based Transformer Inference}
\author{Xiaolong Liang, Juanjuan Li, Rui Qin, and Yisheng Lv\\[0.5em]
\small State Key Laboratory of Multimodal Artificial Intelligence Systems,\\[-0.1em]
\small Institute of Automation, Chinese Academy of Sciences, Beijing 100190, China\\
\small \texttt{yisheng.lv@ia.ac.cn}}
\date{}

\maketitle

\begin{abstract}
Outsourced Transformer inference exposes clients to model
substitution and incomplete execution, while direct replay removes the
computational benefit of delegation. We present GKR-HND, a registered-model
protocol for verifying the polynomial backbone of Homomorphic--Nonhomomorphic
Decomposition Transformers. The retained verifier checks the GKR transcript and
registered-weight openings, but delegates expensive public evaluations to an
assigned computation worker. Assuming an honest retained verifier and
prover--worker non-collusion, the verifier
accepts only when the worker's signed, request-bound response agrees with the
proof claims. Experiments with pretrained HND models validate the proof path and
the delegated public computation without dense-matrix replay.
\end{abstract}


\section{Introduction}

Delegating Transformer inference to edge or cloud services reduces the
computational burden on resource-constrained clients~\cite{zheng2025review,
gong2023edge,he2024llms}. At the same time, clients lose direct control over
which model is deployed and how inference is executed. An untrusted service
provider may therefore substitute a smaller model, alter the registered
weights, or omit part of the computation. Although reputation and payment
mechanisms can discourage such behavior, they cannot certify that a particular
inference instance used the registered model and completed the required
computation~\cite{safetynets2017,zhao2021veriml}. Re-executing the model would
provide such a check, but would eliminate the computational benefit of
outsourcing. We therefore ask whether a client can verify the execution of a
registered Transformer backbone without storing its dense weights or replaying
its matrix products.

Cryptographic proof systems can verify outsourced computation without replaying
it~\cite{ggpr13,benSasson2018stark,spartan2020}. Proving a complete Transformer
remains computationally expensive because encoding its non-polynomial
operations enlarges the arithmetic relation. The Homomorphic--Nonhomomorphic
Decomposition (HND) Transformer~\cite{hnd} reduces this arithmetization burden
by separating a polynomial backbone from compact non-polynomial correction
modules. Its original modular verification procedure, however, still propagates
a full hidden state through $L$ dense blocks and retains $O(Ld^2)$ client work.
We therefore present GKR-HND, which applies
GKR~\cite{gkr,thaler2013} to the registered HND backbone.

GKR-HND combines a block-local GKR proof with delegated public computation. The
retained verifier checks the proof transcript and the openings against the
registered model. It derives the remaining public evaluation queries from the
checked transcript and sends them to an assigned computation worker. The worker
does not receive an expected answer from the prover. Its signed response is
bound to the model, request, block, operation, and transcript. With an honest
retained verifier and prover--worker non-collusion, disagreement causes
rejection, while accepted
block results can be composed through their public boundaries.

The contributions are summarized as follows.
\begin{itemize}
    \item We propose GKR-HND, a method for verifying the outsourced polynomial
          backbone of HND inference without replaying dense products, under a
          threat model with an honest retained verifier and non-colluding prover
          and worker.
    \item We separate retained cryptographic verification from public terminal
          computation and bind the assigned worker's response to each query.
    \item We evaluate the protocol with pretrained HND models and verify the
          delegated proof path without dense-matrix replay.
\end{itemize}

The remainder of this paper is organized as follows.
Section~\ref{sec:related} reviews the background and related work.
Section~\ref{sec:problem} formulates the verification problem and introduces
the threat model. Section~\ref{sec:hnd-gkr} presents the GKR-HND verification
protocol. Section~\ref{sec:complexity} analyzes its security guarantees and
computational costs. Section~\ref{sec:evaluation} reports the experimental
results. Finally, Section~\ref{sec:conclusion} concludes the paper.

\section{Background and Related Work}
\label{sec:related}

The GKR protocol allows a verifier to check the claimed output of a layered
arithmetic circuit without re-executing the full
circuit~\cite{gkr,thaler2013}. Starting from the output claim, GKR applies sum-check
to reduce a claim about one layer to claims about the preceding
layer~\cite{lfkn}. For regular circuits, the verifier work is polylogarithmic in
the layer width, apart from input and wiring evaluations. Streaming refinements
further reduce this overhead for structured
computations~\cite{cormode2011}. This layer-by-layer reduction is well matched to the HND
polynomial backbone. Because the backbone is organized as a sequence of
layered blocks, GKR can reduce each block to random-point claims and thereby
avoid client-side replay of its dense matrix operations.

Succinct arguments offer a different design point. Pinocchio and Groth16 use
preprocessing to obtain compact pairing-based proofs~\cite{pinocchio13,groth16},
whereas Aurora and FRI-based systems use transparent polynomial
testing~\cite{aurora2019,deepfri2020,benSasson2018stark,fri2018}. These constructions
make different tradeoffs among setup, proof size, and verifier cost. The
terminal step of GKR requires a narrower mechanism. Once the reduction reaches
evaluations of registered weight tables, the verifier must authenticate them
against the commitments fixed during model registration. KZG and inner-product
constructions provide this mechanism with different setup and proof-size
tradeoffs~\cite{kzg10,bootle2016,bulletproofs2018}. GKR-HND represents the
terminal check through the abstract interface in~\eqref{eq:pec-interface},
leaving the block reduction independent of a particular commitment backend.

Neural inference introduces a different source of cost. The prover workload
depends not only on the model computation but also on its representation inside
the arithmetic relation. SafetyNets verifies matrix multiplication together
with polynomial activation layers~\cite{safetynets2017}, while VeriML applies
SNARKs to selected training iterations~\cite{zhao2021veriml}. Later work
supports more complex inference. Mystique develops matrix proofs and conversion
protocols for committed neural networks, whereas ZKML compiles practical models
to Halo2 circuits~\cite{mystique2021,zkml2024}. zkLLM and zkGPT specialize the
proof procedure for language-model inference~\cite{zkllm2024,zkgpt2025}.
Although polynomial approximations and lookup arguments make Transformer
operations such as softmax and normalization provable, the prover must still
evaluate the encoded relation~\cite{hao2024,zkml2024,zkllm2024}. The
representation itself therefore contributes to the proof cost.

Model design can reduce this cost before proof generation. Quantization maps
real-valued computation to low-precision
arithmetic~\cite{hubara2018quantized,jacob2018quantization}, while polynomial
networks replace nonlinear activations with algebraic
surrogates~\cite{ivakhnenko1971polynomial}. Several Transformer variants likewise change
the computational structure of attention~\cite{katharopoulos2020linear,
wang2020linformer,kitaev2020reformer,choromanski2021performer}. HND addresses
the problem through decomposition. It separates inference into a polynomial
backbone and compact non-polynomial correction modules~\cite{hnd}. Because the
backbone is already a layered polynomial computation, its verification can use
the correction values as fixed public inputs instead of placing the correction
modules inside the same relation. This reduces the relation that must be
proved. The original HND audit, however, still propagates a full hidden state
through $L$ dense blocks and retains $O(Ld^2)$ client work. The verification
bottleneck therefore remains.

Verification can also separate cryptographic checking from expensive public
computation. Flow provides a systems precedent for assigning execution and
result checking to different node roles~\cite{flow2019verification}. GKR-HND
uses a narrower separation. The retained verifier checks the GKR transcript and
registered openings, while an assigned worker evaluates the public tables at
the verifier-derived points. The worker signs its response, and any mismatch
causes rejection. This arrangement removes dense replay from the client without
transferring the verifier's sum-check decisions to the worker.

\section{Problem Formulation}
\label{sec:problem}

\subsection{Problem Statement}

We consider a registered public HND model over $\Fp$. Before any inference
request, a model publisher submits the architecture and weights
$W=\{W_l\}_{l=0}^{L-1}$ to a trusted registrar. The registrar validates their
declared structure and numeric bounds, fixes a commitment
$\mathsf{cm}_{W_l}$ to each encoded weight table, and publishes the resulting
model identifier. At inference time, the client delegates backbone execution to
an untrusted prover $P$. A retained verifier $V$ in the client trust domain
checks the GKR transcript and registered-weight openings, while independently
operated workers may perform designated public computations for different
blocks. The public statement for the proved backbone is
\begin{equation}
  \mathsf{stmt}=(\mathsf{modelID},\mathsf{requestID},X_0,C,X_L),
  \label{eq:public-statement}
\end{equation}
where $X_0$ is the encoded input, $C=\{C_l\}_{l=0}^{L-1}$ contains correction
values fixed before proof challenges, $X_L$ is the claimed backbone output, and
$\mathsf{requestID}$ binds the encoded input, correction tables, and typed
initial boundary selected for this inference to the registered backbone and
block. It also binds the token count, numeric and output policies, assigned
worker and role specification, parameter registry, and a freshness nonce. Raw
input tokens and embedding parameters remain outside this descriptor because
the verified relation begins at $X_0$. A prover-side helper $A$ may generate
weight-evaluation proofs and is treated as part of the prover side.

For compact notation,
\begin{equation*}
 W_l=(W_Q,W_K,W_V,W_O,W_1,W_2,\gamma_1,\gamma_2)
\end{equation*}
denotes the tuple of block-$l$ tensors, and
$\mathsf{cm}_{W_l}$ denotes the corresponding tuple of eight individual
commitments. Model dimensions and the canonical zero-padding rule are included
in $\mathsf{modelID}$.

The cryptographic verification problem is to decide whether the claimed
backbone output matches the registered computation:
\begin{equation}
\begin{aligned}
    X_L \stackrel{?}{=}\;&
    H_{L-1}(\cdots H_1(H_0(X_0;W_0,C_0);W_1,C_1)\cdots; \\
    &\qquad W_{L-1},C_{L-1}).
\end{aligned}
\label{eq:hnd-verification-target}
\end{equation}
The retained verifier checks this statement without replaying the full
backbone. For a weight table $W$, we require a polynomial-evaluation commitment
interface
\begin{align}
 \mathsf{cm}_{W}&\leftarrow\mathsf{Commit}(W), \notag\\
 (v,\pi_W)&\leftarrow\mathsf{Open}(W,r), \notag\\
 \{0,1\}&\leftarrow
 \mathsf{VerifyEval}(\mathsf{cm}_{W},r,v,\pi_W),
 \label{eq:pec-interface}
\end{align}
where an accepted opening is intended to establish
$v=\widetilde W(r)$. The backend may be KZG-, inner-product-, or
hash/folding-based; GKR-HND uses only this interface.

We assume evaluation soundness for this interface. Once a commitment is fixed
and before $r$ is sampled, a probabilistic polynomial-time adversary can produce
an accepted tuple $(r,v,\pi_W)$ with $v\ne\widetilde W(r)$ only with probability
at most $\epsilon_{\mathrm{eval}}$. The deployment-level verification task also
contains deterministic checks outside this relation. The retained verifier may
assign the public fixed-point scan and terminal public evaluations to an
independent worker. Combining the returned values with the proved backbone
yields a hybrid protocol: GKR covers the polynomial relation, while the worker
independently evaluates the assigned public computation.

\subsection{Threat Model}

The registrar is trusted to validate the declared model and to bind the
canonical encoding, numeric policy, and public parameters to
$\mathsf{modelID}$. The client $C$ retains verifier $V$ as the acceptance
authority in its trust domain. Verifier $V$ fixes the request, checks the sum-check transcript in
order, derives the Fiat--Shamir challenges, verifies the ML--KZG openings, and
composes the accepted block boundaries. Corruption of $V$ is therefore outside
the integrity model.

The prover $P$ executes the registered backbone. Its optional opening helper
$A$ belongs to the same trust domain and may be controlled by the same
adversary. This arrangement leaves the registered-weight claims protected by
evaluation soundness because $V$ verifies each opening against the pinned
commitment.

For block $l$, verifier $V$ may assign the expensive public-tensor evaluations
and fixed-point scan to a worker $O_l$. The
worker receives evaluation points derived by $V$ after the corresponding
prover messages have been fixed, and returns a result bound to the model,
request, block, transcript, and assigned operation. Verifier $V$ compares that
result with the proof-bound values and rejects a disagreement. For a delegated
check outside the cryptographic relation, integrity assumes that at least one
of $P$ and $O_l$ follows the prescribed computation and that they are not
jointly controlled. Different blocks may use different workers and proceed
concurrently; each block requires one assigned result. The artifact records
this role convention in \texttt{docs/protocol\_roles.json} under the protocol
identifier \texttt{GKR-HND/v1}.

\subsection{Evaluation Commitment Backend}

The implementation instantiates \eqref{eq:pec-interface} with a direct
multilinear extension of the KZG evaluation paradigm~\cite{kzg10}. We specify
it here because its setup and opening costs are part of the measured system.
Let $(\mathbb G_1,\mathbb G_2,\mathbb G_T,e)$ be the asymmetric pairing groups
of BLS12-381, write $[z]_k=zG_k$, and let
$\chi_b(x)=\prod_{i=0}^{m-1}x_i^{b_i}(1-x_i)^{1-b_i}$ for
$b\in\{0,1\}^m$. Setup samples $\tau=(\tau_0,\ldots,\tau_{m-1})$ and publishes
\begin{equation}
 \begin{split}
 \mathsf{pp}_m=\big(&G_1,G_2,\{[\tau_j]_2\}_{j=0}^{m-1},
 \{[\chi_b(\tau)]_1\}_{b\in\{0,1\}^m},\\
 &\{[\chi_b(\tau_{j+1:m})]_1:
 j=0,\ldots,m-1,\ b\in\{0,1\}^{m-j-1}\}\big).
 \end{split}
 \label{eq:mlkzg-srs}
\end{equation}
The setup-local scalars $\tau$ are erased. The full Lagrange basis permits a
commitment directly from an LSB-first evaluation table:
\begin{equation}
 \mathsf{cm}_f=\sum_{b\in\{0,1\}^m}f(b)[\chi_b(\tau)]_1=[f(\tau)]_1.
 \label{eq:mlkzg-commit}
\end{equation}

To open at $r=(r_0,\ldots,r_{m-1})$, define one suffix quotient per variable,
\begin{equation}
q_j(x_{>j})=
 f(r_{<j},1,x_{>j})-f(r_{<j},0,x_{>j}),
 \label{eq:mlkzg-quotient}
\end{equation}
and return $v=f(r)$ together with
$Q_j=[q_j(\tau_{>j})]_1$. Successive table folding constructs these
quotients without coefficient conversion. Multilinearity gives the telescoping
identity
\begin{equation}
 f(\tau)-f(r)=\sum_{j=0}^{m-1}(\tau_j-r_j)q_j(\tau_{>j}),
 \label{eq:mlkzg-telescope}
\end{equation}
so verification checks the single multi-pairing equation
\begin{equation}
 e(\mathsf{cm}_f-[v]_1,G_2)
 \prod_{j=0}^{m-1}e(-Q_j,[\tau_j-r_j]_2)=1_{\mathbb G_T}.
 \label{eq:mlkzg-verify}
\end{equation}

For an $m$-variate table, this experimental SRS stores
$2^{m+1}-1$ G1 elements, in addition to $m$ G2 trapdoor encodings. Commitment
and opening are $O(2^m)$ and the proof contains one field value and $m$ G1
elements (32$+48m$ compressed bytes on BLS12-381); verification uses one
multi-pairing with $m+1$ terms. Every public parameter is canonically serialized
and hashed into the registered SRS digest. The implementation does not batch
openings and this paper does not claim a new standalone security reduction for
the backend. The analysis in Section~\ref{sec:soundness} represents its
contribution by $\epsilon_{\mathrm{eval}}$. Production use should replace it
with an audited PCS that satisfies the evaluation-soundness condition above
under a stated setup ceremony and assumption.

The protocol provides execution integrity under this assumption and fresh
verifier challenges. It does not provide model or input confidentiality,
availability, correct model registration, or semantic model quality. A failed
or withholding prover-side opening helper can delay verification, but it cannot make an
invalid execution pass unless it breaks evaluation soundness.

\subsection{HND Decomposition}

The HND architecture describes a network through two coordinated components:
\begin{equation}
    f(x) \approx H(x,\,N(x)),
    \label{eq:hnd-decomposition}
\end{equation}
where $H$ is a polynomial backbone over $\Fp$ and $N$ supplies correction
values for non-polynomial operations. The verification relation studied here is
\emph{conditional}: it proves $H(X_0,C)$ for correction values $C$ already
included in the public statement. It does not prove that $C=N(X)$ or that an
approximate correction generator preserves the accuracy of an undecomposed
Transformer. This distinction is necessary because a verifier that does
not materialize intermediate states cannot in general compute state-dependent
normalization factors for free.

\begin{figure}[htbp]
\centering
\includegraphics[width=0.58\linewidth]{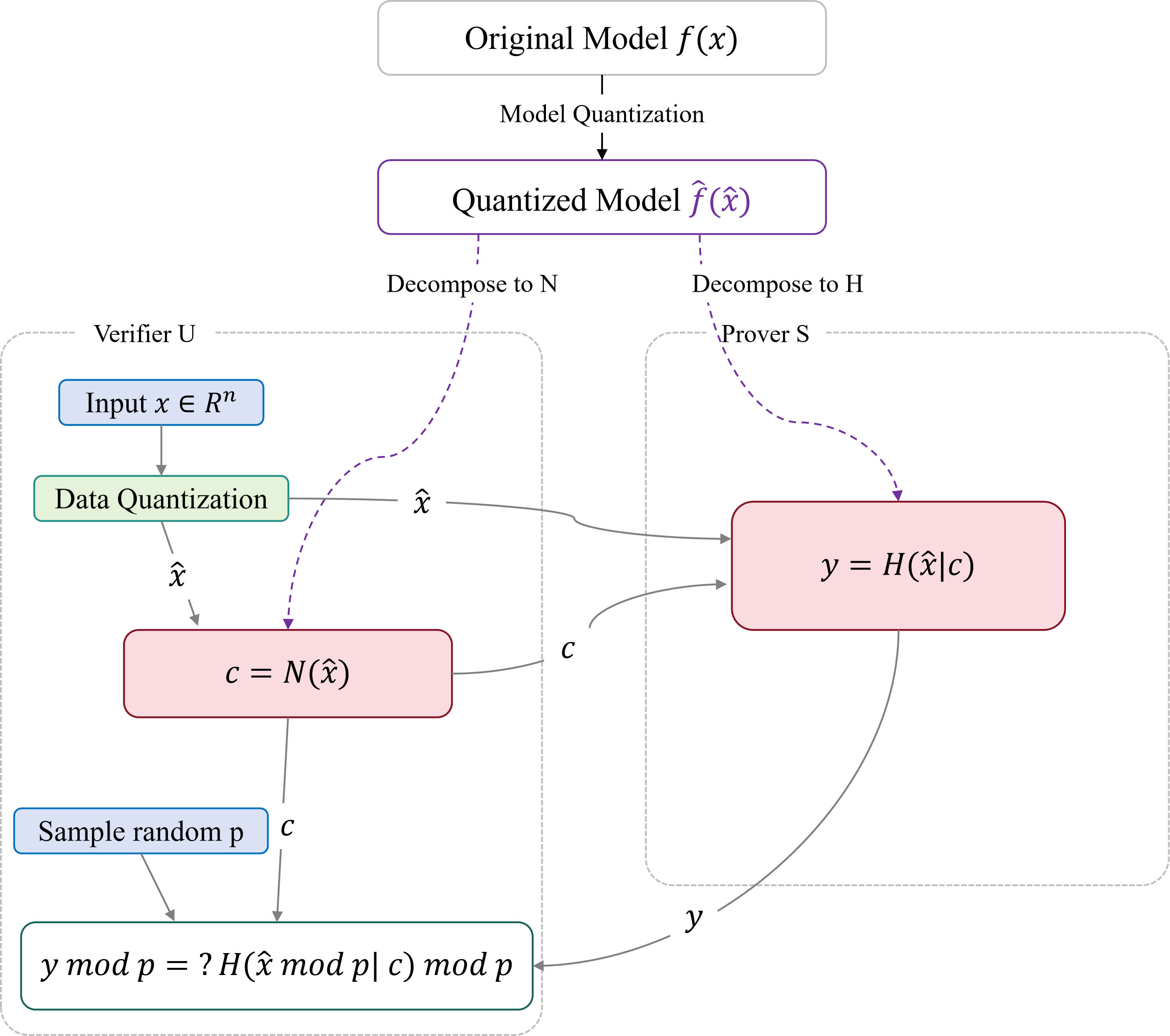}
\caption{Overview of the original HND decomposition. The quantized model is
separated into a compact correction network $N$ and a polynomial backbone $H$.
The intended deployment evaluates the correction path on the verifier side and
the backbone on the prover side. In the checkpoint experiment in this paper,
the resulting correction tables are fixed as public inputs; their generation
is not part of the proved relation.}
\label{fig:hnd-decomposition}
\end{figure}

In the ideal HND architecture, the correction would satisfy
$C_l^\star=N_l(X_l)$. The verified relation instead treats $C_l$ as a public
input and checks only
\begin{equation}
    X_{l+1} = H_l(X_l; W_l, C_l) \in \Fp^{T\times d}.
    \label{eq:hnd-layer-target}
\end{equation}
Here $W_l$ is registered before the request, $C_l$ is public and fixed before
the challenges, and $H_l$ is the outsourced polynomial computation. An
extended hybrid protocol can assign the deterministic check $C_l=N_l(X_l)$ to
a computation worker without placing it in the GKR relation. The current
checkpoint proof experiment does not yet implement that extension. It imports
correction tables produced by the original networks and fixes them as public
inputs. The separate scaling benchmark uses nonzero synthetic field elements
to exercise the polynomial kernel. Neither experiment alone constitutes
end-to-end verified language-model inference.

The core Transformer components fit this structure as follows. RMSNorm contains
a reciprocal square-root scaling factor. HND moves this scalar into the
correction network and leaves the backbone with a multiplication by a trusted
scale:
\begin{equation}
    \mathrm{RMSNorm}(x)=\gamma\cdot x\cdot
    \frac{1}{\sqrt{\frac{1}{d}\sum_i x_i^2+\epsilon}}
    \quad\leadsto\quad
    \gamma\cdot x\cdot c_{\mathrm{rms}} .
\end{equation}
Softmax attention is replaced by a polynomial attention-like block, such as a
SimA-style construction. With
\begin{equation}
    Q=XW_Q,\qquad K=XW_K,\qquad V=XW_V,
\end{equation}
the non-polynomial normalization factors are isolated as correction scalars,
for example
\begin{align}
    \widehat Q &= Q\odot C_q, &
    \widehat K &= K\odot C_k, \notag \\
    S &= \widehat Q\widehat K^\top, &
    O &= SV.
\end{align}
The server-side operations are then matrix multiplication and element-wise
multiplication over $\Fp$. Smooth activations are replaced by polynomial
surrogates, for example
\begin{equation}
    \phi(z)=az^2+bz+c,
\end{equation}
or by wider polynomial feed-forward blocks. Residual additions remain native
field additions; the evaluated checkpoints use $c=0$. In the HND Transformer
instantiation, the correction table contains RMSNorm inverse scales and Q/K
L1-inverse attention scales for every token, i.e., $T(2H+2)$ scalars per layer
for sequence length $T$ and $H$ heads.

\subsection{The GKR Protocol}

GKR is an interactive proof for layered arithmetic computations.
In the HND setting, each backbone layer is viewed as a finite-field arithmetic
layer $V_i\mapsto V_{i+1}$, where gate values correspond to hidden-state
coordinates. The verifier starts from a claim about the output layer and
repeatedly reduces it to a claim about the previous layer. Each reduction
invokes the sum-check protocol, which is the core per-layer
mechanism.

Consider a low-degree polynomial $f(x_1,\dots,x_v)$ over $\Fp$. The prover
claims a value $H$ for the sum over all Boolean assignments:
\begin{equation}
    H = \sum_{x_1 \in \{0,1\}}\!\cdots\!\sum_{x_v \in \{0,1\}} f(x_1,\dots,x_v).
\end{equation}
Sum-check verifies this claim in $v$ rounds without enumerating the $2^v$
Boolean assignments. In round $j$, the prover sends the univariate polynomial
$g_j(t)=\sum_{x_{j+1},\dots,x_v\in\{0,1\}} f(r_1,\dots,r_{j-1},t,x_{j+1},\dots,x_v)$,
where $r_1,\dots,r_{j-1}$ are random challenges from earlier rounds. If the
individual degree in round $j$ is bounded by $\delta_j$, then $g_j$ is sent
using $\delta_j+1$ coefficients. The verifier checks
$g_j(0)+g_j(1)=g_{j-1}(r_{j-1})$ (with $g_0(r_0)\triangleq H$) and samples a
fresh challenge $r_j\in\Fp$. After $v$ rounds, the verifier accepts the
original claim if and only if $f(r_1,\dots,r_v)=g_v(r_v)$. By the
Schwartz--Zippel bound, a cheating prover succeeds with probability at most
$\sum_{j=1}^v \delta_j/|\Fp|$ for this sum-check instance. Multilinear
relations have $\delta_j=1$, while product gates and polynomial activations
use the corresponding constant degree.

Direct evaluation of $f(r_1,\dots,r_v)$ would be as expensive as recomputing
the layer. GKR avoids this by encoding the previous layer $V_{i-1}$ as $f$,
so that the claim $f(r_1,\dots,r_v)=?$ becomes the input to the next
recursion step, pushing the verification burden down layer by layer:
\begin{equation}
    \text{claim about } V_L
    \Longrightarrow
    \text{claim about } V_{L-1}
    \Longrightarrow \cdots \Longrightarrow
    \text{claim about } V_0 .
\end{equation}
The verifier never materializes intermediate layers. The prover supplies the
low-degree univariate polynomials and a small number of random-point
evaluations, and the final claim reduces to the known input $V_0=X_0$. In a
non-interactive deployment, the verifier's random challenges can be derived
from an auditable public transcript.

Soundness requires each challenge to be unpredictable when the prover fixes the
message it checks. In the interactive protocol, $V$ samples a fresh challenge
after receiving each sum-check polynomial. The public statement in
\eqref{eq:public-statement} and the model commitments are fixed before the first
challenge. Any invalid layer relation then induces a nonzero low-degree
discrepancy, which a fresh random point detects except with the stated
Schwartz--Zippel probability. Weight claims are not accepted from an oracle:
they must pass \eqref{eq:pec-interface} against the preregistered commitments.

A non-interactive transformation is possible only when every challenge is
derived from the entire preceding transcript. For example,
\begin{equation*}
 r_j=H\!\left(\mathsf{domain},\mathsf{stmt},
 \{\mathsf{cm}_{W_l}\},m_1,r_1,\ldots,m_j\right),
\end{equation*}
where $m_j$ is the prover message preceding $r_j$. A randomness beacon may be
included as a salt, but cannot replace this sequential binding. Security of
that transformation is in the random-oracle model. The non-interactive
prototype in Section~\ref{sec:evaluation} implements the sequential transcript
binding, but its experiments do not measure ROM security.

\section{GKR-HND Verification Protocol}
\label{sec:hnd-gkr}

\begin{figure}[htbp]
\centering
\includegraphics[width=\linewidth]{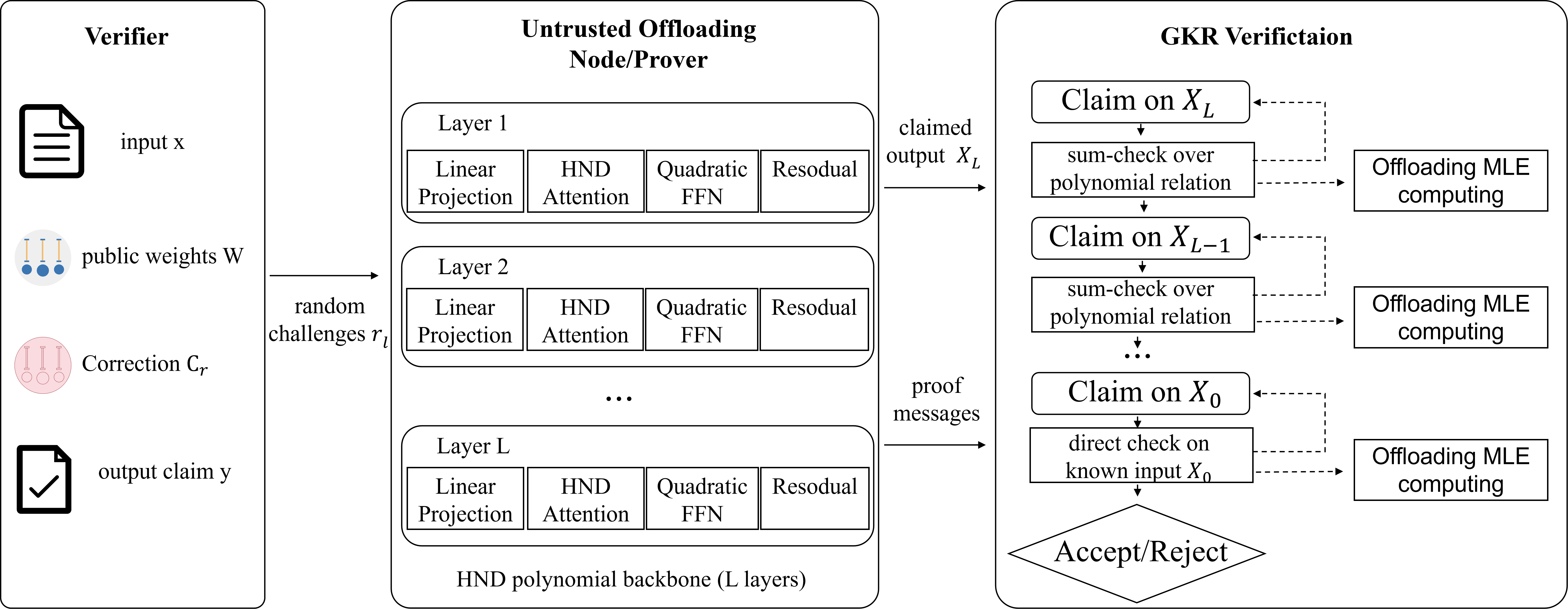}
\caption{Relation-level GKR-HND workflow used by the retained verifier. The
diagram illustrates backward reduction across layered polynomial computation;
the deployment instantiates this check for each registered block.
The ``public weights'' label denotes the conceptual interface, which the
prototype realizes with pinned commitments and ML--KZG openings rather than
dense-weight replay. The block-local interface supports later authenticated
composition across the registered backbone.}
\label{fig:gkr-hnd-workflow}
\end{figure}

This section describes the two layers of GKR-HND verification. At the relation
layer, the prover executes the polynomial backbone and produces one registered
proof artifact per block. At the delegation layer, the retained verifier checks
the proof and derives queries for the public computation assigned to a separate
worker. The worker evaluates the disclosed data and signs each response. The
backbone trace is represented as a layered arithmetic computation over $\Fp$:
\begin{equation}
    X_{l+1}=H_l(X_l;W_l,C_l),\qquad l=0,\ldots,L-1,
    \label{eq:hnd-trace}
\end{equation}
where $W_l$ is registered and $C_l$ is fixed in the public statement. The
retained verifier checks the transition from $X_l$ to $X_{l+1}$ without
replaying its dense matrix products, as shown in
Fig.~\ref{fig:gkr-hnd-workflow}. Block artifacts share public boundary tensors,
so the client can compose accepted transitions without constructing a single
high-degree, all-block relation.

\subsection{Delegated Public Computation and Result Composition}

The client first fixes a request root
\begin{equation}
 \mathsf{R}_{\mathrm{req}}=
 \mathsf{Hash}(\mathsf{domain},\mathsf{descriptor}_{\mathrm{canonical}}),
 \label{eq:request-root}
\end{equation}
where the canonical descriptor contains a nonzero fresh nonce and binds the
registered backbone and block, assigned worker and role specification, encoded
input and typed initial boundary, correction data, token count, numeric and
output policies, and parameter-registry digest. The prover then
publishes, for every block $l$, its public input and output boundaries,
correction tables, fixed-point witness $\mathcal W_l$, and proof $\pi_l$. These
artifacts are delivered to the retained verifier; the worker receives only the
registered public data and the queries derived by that verifier.

The retained verifier $V$ authenticates the block registration and checks
$\pi_l$ against its public boundaries and registered commitments. Only after a
sum-check reduction succeeds does $V$ issue a terminal query. The query binds
the protocol version, model and request identifiers, block and operation
indices, transcript digest, evaluation point, numeric policy, and freshness
nonce. The assigned worker $O_l$ scans $\mathcal W_l$ and evaluates the
requested public tables without receiving an expected value from the prover.

For query $Q_{l,j}$ with binding metadata $B_{l,j}$, the worker returns a
canonical response
\begin{equation}
 S_{l,j}=(B_{l,j},\mathsf{Hash}(Q_{l,j}),\mathsf{id}_{O_l},
 \mathsf{status}_{l,j},\mathsf{error}_{l,j},\mathsf{value}_{l,j},
 \mathsf{vk}_{O_l},\sigma_{l,j}),
 \label{eq:worker-response}
\end{equation}
where $\sigma_{l,j}$ is an Ed25519 signature under the public key assigned to
$O_l$. The verifier checks the key, signature, complete query binding, and
returned value before continuing the transcript. A missing, duplicated, stale,
malformed, or inconsistent response causes rejection.

The retained verifier accepts block $l$ only after the proof, registered
openings, numeric scan, and all delegated responses succeed. It accepts the
backbone chain after every required block is present in order and the output
digest of block $l$ equals the input digest of block $l+1$. The proof relation
remains block-local; chain composition compares authenticated block results and
does not form one all-block polynomial.

The same block decisions can be accumulated without changing the per-block proof
relation. For example, a history value suitable for a later PoH-style extension
can be updated as
\begin{equation}
 h_{l+1}=\mathsf{Hash}(h_l,\mathsf{R}_{\mathrm{req}},l,B_l,B_{l+1},
 A_l,\{\sigma_{l,j}\}_j),
 \label{eq:response-history}
\end{equation}
where $\{\sigma_{l,j}\}_j$ are the worker responses used for block $l$. This hash
chain records the ordered sequence of accepted transitions for a future
PoH-style extension while each GKR proof remains block-local.

\subsection{Tensor Encoding and Multilinear Extensions}

We represent every HND backbone tensor by its multilinear extension (MLE) over
a Boolean hypercube. For a hidden tensor of shape $T\times d$, choose
\begin{equation*}
 m_T=\lceil\log_2 T\rceil,\qquad m_d=\lceil\log_2 d\rceil,
\end{equation*}
pad with zeros, and index entries by
$(t,i)\in\{0,1\}^{m_T}\times\{0,1\}^{m_d}$. Let
$\Hcube=\{0,1\}^m$. For $i=(i_1,\ldots,i_m)\in\Hcube$, the Lagrange basis
polynomial is
\begin{equation}
    \beta_i(x)=\prod_{k=1}^{m}\big(x_k i_k+(1-x_k)(1-i_k)\big),
    \qquad x\in\Fp^m.
\end{equation}
For an array $A$ indexed by $\Hcube$, its multilinear extension is
\begin{equation}
    \MLE{A}(x)=\sum_{i\in\Hcube}A_i\,\beta_i(x).
    \label{eq:mle-def}
\end{equation}
For matrices, we use one group of variables per index:
\begin{equation}
    \MLE{W}(r,s)=\sum_{i,j}W_{i,j}\,\beta_i(r)\,\beta_j(s).
\end{equation}

A random evaluation $\MLE{A}(r)$ acts as a fingerprint of the entire tensor:
changing any coordinate of $A$ changes the polynomial $\MLE{A}$, and a random
evaluation detects the discrepancy except with probability bounded by the
Schwartz--Zippel bound. This is why GKR works with multilinear extensions
rather than ordinary array indices. However, computing $\MLE{A}(r)$ directly
still requires summing over all $2^m$ Boolean assignments. The saving comes
from sum-check: the prover handles the large hypercube sums, while the
verifier checks only low-degree univariate identities and samples fresh
challenges. For a layer of width $d=2^m$, the verifier performs
$O(m)=O(\log d)$ such checks rather than enumerating all $d$ coordinates.
Layer by layer, GKR turns a random-point claim about $X_{l+1}$ into a
random-point claim about $X_l$, until the final claim concerns the known
input $X_0$.

\subsection{Registered HND Block Relation}

The verified backbone relation is structurally aligned with the pre-norm
computation used by the available HND checkpoints. Let
$X_l\in\Fp^{T\times d}$, $d=Hd_h$, and let
$M_{t,u}=1$ when $u\le t$ and zero otherwise. The client-supplied correction
tables are $C_{q,l},C_{k,l}\in\Fp^{T\times H}$ and
$C_{1,l},C_{2,l}\in\Fp^T$. For every block $l$:
\begin{equation}
\begin{aligned}
    H^{(1)}_l[t,j] &= X_l[t,j]C_{1,l}[t]\gamma_{1,l}[j],\\
    Q_l&=H^{(1)}_lW_{Q,l},\quad
    K_l=H^{(1)}_lW_{K,l},\quad
    V_l=H^{(1)}_lW_{V,l},\\
    D_{l,h}[t,u]&=\sum_{j=0}^{d_h-1}
       Q_l[t,hd_h+j]K_l[u,hd_h+j],\\
    S_{l,h}[t,u]&=M_{t,u}D_{l,h}[t,u]
       C_{q,l}[t,h]C_{k,l}[u,h],\\
    Z_{l,h}[t,j]&=\sum_{u=0}^{T-1}S_{l,h}[t,u]
       V_l[u,hd_h+j],\\
    Y_l^{\mathrm{attn}}&=\operatorname{Concat}_h(Z_{l,h})W_{O,l},\qquad
    U_l=X_l+Y_l^{\mathrm{attn}},\\
    H^{(2)}_l[t,j]&=U_l[t,j]C_{2,l}[t]\gamma_{2,l}[j],\\
    G_l&=H^{(2)}_lW_{1,l},\qquad
    P_l=a_lG_l^{\odot2}+b_lG_l,\\
    F_l&=P_lW_{2,l},\qquad X_{l+1}=U_l+F_l.
\end{aligned}
\label{eq:fw-block}
\end{equation}
The six matrices and two normalization vectors are bound by ML--KZG
commitments; $(a_l,b_l)$, the dimensions, padding convention, and SRS digest
are included in the registered model identifier. Corrections are fixed as
public statement data before Fiat--Shamir challenges. The current proof
boundary begins at the embedded backbone input $X_0$ and ends at $X_L$; it
does not yet include token/position embedding lookup, the correction generator,
final RMSNorm, the LM head, sampling, or a key--value cache. The supporting
perplexity audit is therefore a checkpoint-stability diagnostic, not a proved
output.

\subsection{Deterministic Fixed-Point Target Semantics}

To make the remaining checkpoint-to-field gap testable rather than implicit,
we fix the intended integer semantics. For scale $s\in\mathbb{Z}_{>0}$, an
FP32 value $x$ is exported as
\begin{equation}
 q_s(x)=\operatorname{round}_{\mathrm{away}}(sx),
 \label{eq:quantize}
\end{equation}
where nearest rounding is used and exact half cases move away from zero. A
signed integer $z$ is encoded as $z\bmod p$, subject to a public magnitude
bound $|z|\le B<p/2$ so that the field representation is unambiguous. After a
fixed-point product or dot-product accumulator $A$, rescaling uses witnesses
$(Q,R)$ satisfying
\begin{equation}
 A=sQ+R,\qquad 2|R|\le s,
 \label{eq:rescale-witness}
\end{equation}
with the sign of an exact-half remainder chosen opposite to that of $A$. This
uniquely implements the rounding rule in \eqref{eq:quantize}.

For the imported block, $\operatorname{RS}(A)$ denotes the quotient $Q$ in
\eqref{eq:rescale-witness}. The executable source-order schedule is
\begin{equation}
\begin{aligned}
 T_1&=\operatorname{RS}(X C_1),& H_1&=\operatorname{RS}(T_1\gamma_1),\\
 Q&=\operatorname{RS}(H_1W_Q),&K&=\operatorname{RS}(H_1W_K),&
 V&=\operatorname{RS}(H_1W_V),\\
 Q'&=\operatorname{RS}(Q C_q),&K'&=\operatorname{RS}(K C_k),\\
 D_h&=\operatorname{RS}(Q'_hK_h^{\prime\top}),&S_h&=M\odot D_h,\\
 Z_h&=\operatorname{RS}(S_hV_h),&O&=\operatorname{RS}(\operatorname{Concat}_h Z_h W_O),\\
 U&=X+O,&T_2&=\operatorname{RS}(U C_2),&H_2&=\operatorname{RS}(T_2\gamma_2),\\
 G&=\operatorname{RS}(H_2W_1),&A_G&=\operatorname{RS}(aG),\\
 Q_G&=\operatorname{RS}(A_G\odot G),&B_G&=\operatorname{RS}(bG),&P&=Q_G+B_G,\\
 F&=\operatorname{RS}(PW_2),&X^+&=U+F.
\end{aligned}
\label{eq:fixed-schedule}
\end{equation}
This ordering follows the PyTorch source; it is intentionally not rearranged
using real-number associativity because intermediate rounding would change.

The fixed-point mode publishes every quotient and remainder table. Before any
matrix claim is accepted, the verifier linearly checks tensor shapes, the
causal mask, both residual additions, every pointwise product, and the exact
half-away rule in \eqref{eq:rescale-witness}. Every public signed value obeys
$|z|\le2^{62}-1$. For each matrix product, a conservative
$n\max|A|\max|B|$ bound and the reconstructed accumulator $sQ+R$ must fit
signed 128-bit arithmetic and remain below $p/2$. A field equality established
by sum-check therefore lifts to the unique signed integer equality without
modular wraparound. Six checkpoint-weight products use ML--KZG openings; the
$H$ score and $H$ value-aggregation products use public-table terminal
evaluations, for $6+2H=22$ matrix sum-checks at $H=8$.

This design closes the numerical-semantics gap for the disclosed public block
witnesses, but it is not a succinct range proof: the numeric scan costs linear
time in the 2.08-MB public-instance encoding. It avoids replaying dense matrix
multiplications, not reading the public witness. The offline registrar imports
the SHA-256-pinned Q-format export and recomputes the canonical block
registration, including all eight commitments. It signs the registration only
if the recomputed bytes match the submitted statement. The resulting model
identifier binds the source-checkpoint label and block position to the
quantized weights, public model metadata, numeric and padding rules, and SRS.
This procedure authenticates the exported fixed-point model; it does not prove
semantic equivalence to the FP32 checkpoint. Per inference, the retained
verifier receives the signed registration, public witness, proof, and public
SRS. The computation worker receives the signed registration, public witness,
and verifier-derived evaluation query, but not the proof. The 25-MB exporter
file and dense weights are absent from both online interfaces. We claim no
coverage of correction generation or the full checkpoint.

Each nontrivial check in the backward pass is either a sum-check instance or a
low-degree consistency check at an induced random point. For a matrix product
$Y=XW$ with contracted dimension $b$, the exact MLE identity is
\begin{equation}
 \widetilde Y(r_x,r_y)=
 \sum_{j\in\{0,1\}^{\lceil\log_2 b\rceil}}
 \widetilde X(r_x,j)\widetilde W(j,r_y),
 \label{eq:matmul-mle}
\end{equation}
where every padded coordinate is constrained to zero. The implementation stores
the full padded evaluation tables: public inputs and registered tensors are
rejected unless canonical, unused weight entries are fixed to zero, and the
operator relations propagate those zeros through intermediate tables. There is
no separate validity-mask argument. Sum-check over $j$
ends in one claim on $X$ and one evaluation claim on $W$. A pointwise product
$Y=A\odot B$ uses
\begin{equation}
 \widetilde Y(r)=\sum_{i\in\{0,1\}^{m}}
 \beta_i(r)\widetilde A(i)\widetilde B(i),
 \label{eq:pointwise-mle}
\end{equation}
and addition is checked by MLE linearity. The head selectors and concatenation
in \eqref{eq:fw-block} are fixed wiring predicates. More generally, a
contracted operator has the form
\begin{equation}
    \MLE{Y}(r)=\sum_{z\in\{0,1\}^{m}} F(z;r),
    \label{eq:generic-op}
\end{equation}
where $F$ is the explicit low-degree integrand induced by
\eqref{eq:matmul-mle} or \eqref{eq:pointwise-mle}. Its terminal evaluation
yields random-point claims about the operator inputs. Hidden-state claims are
carried backward; every static-weight claim is resolved through
$\mathsf{VerifyEval}$, not by trusting the supplied value. For
\eqref{eq:fw-block}, one block requires claims on six matrices and two
normalization vectors before optional batching.

When the backward pass produces multiple claims about the same tensor, they
are combined by equality-polynomial folding. Suppose the verifier holds two
claims $\MLE{A}(r_1)=a_1$ and $\MLE{A}(r_2)=a_2$. It samples
$\eta\in\Fp$, forms $a_\eta=\eta a_1+a_2$, and rewrites the combined claim
as
\begin{equation}
    a_\eta=
    \sum_{u\in\{0,1\}^m} A_u\big(\eta\beta_u(r_1)+\beta_u(r_2)\big).
    \label{eq:claim-folding}
\end{equation}
Applying sum-check to \eqref{eq:claim-folding} yields a terminal value
$\lambda_\eta(\rho)\MLE{A}(\rho)$, where
$\lambda_\eta(\rho)=\eta\beta_\rho(r_1)+\beta_\rho(r_2)$. If
$\lambda_\eta(\rho)\ne0$, the verifier obtains one claim by division. The
implementation absorbs a unique fold-counter value of zero and accepts no
prover-selected retry, preventing Fiat--Shamir grinding. If the terminal
coefficient is zero, proving fails closed. If either original claim is false,
a fresh $\eta$ cancels its error with probability at most $1/|\Fp|$. The
procedure extends to more claims by repeated folding.

Starting from a random claim on $X_{l+1}$, the verifier first splits
$X_{l+1}=U_l+F_l$, reduces $F_l=P_lW_{2,l}$, checks
$P_l=a_lG_l^{\odot2}+b_lG_l$, and reduces
$G_l=H_l^{(2)}W_{1,l}$.  A degree-four product sum-check verifies
$H_l^{(2)}=U_l\odot C_{2,l}\odot\gamma_{2,l}$, after which the two claims on
$U_l$ are folded. The verifier then splits
$U_l=X_l+Y_l^{\mathrm{attn}}$ and reduces
$Y_l^{\mathrm{attn}}=Z_lW_{O,l}$.

The column variables of $Z_l$ are decomposed into within-head and head-index
bits. For every $h$, a bilinear sum-check reduces
$Z_{l,h}=S_{l,h}V_{l,h}$ over the key-token index. A pointwise degree-three
check enforces the fixed causal mask, and a degree-four check enforces the
query/key correction factors. The score claim is then reduced through
$D_{l,h}=Q_{l,h}K_{l,h}^{\mathsf T}$. Claims from all heads are folded into one
claim on each of $Q_l$, $K_l$, and $V_l$, followed by the three committed
matrix reductions from $H_l^{(1)}$. The resulting claims on $H_l^{(1)}$ are
folded, the first pre-normalization is checked, and its $X_l$ claim is folded
with the attention-residual claim. This message order fixes every sum-check
polynomial before its associated challenge.

The backward reduction is instantiated independently for every block. The
retained verifier fixes the public output $X_{l+1}$ before sampling its first
challenge and reduces the resulting claim to $X_l$. It checks every sum-check
equation and registered-weight opening before deriving any delegated terminal
query. The assigned computation worker evaluates that query over the public
boundary or trace and returns a signed response. The verifier accepts the local
proof only when the response matches the reduced claim. A chain verifier then
composes the block-local decisions by exact equality of adjacent public
boundaries.

\section{Protocol Guarantees and Cost}
\label{sec:complexity}
This section separates the algebraic soundness of each registered block proof
from the deterministic public computation delegated to the assigned worker.
Their combination is analyzed with an honest retained verifier and
prover--worker non-collusion.

\subsection{Time and Space Complexity}

The retained verifier checks every sum-check message and verifies the
experimental ML--KZG openings. The assigned worker scans the fixed-point
witness and evaluates the requested public boundary or trace table. Let
$m_{\max}=\lceil\log_2\max(T,d,d_{\mathrm{ff}})\rceil$. For block $l$, let
$q_{\mathrm{sc},l}$ and $q_{\mathrm{eval},l}$ be the numbers of sum-checks and
terminal weight claims, and let $N_{\mathrm{pub},l}$ be the number of
disclosed numeric entries. Their combined online work is
\begin{equation}
\begin{aligned}
 &O(|X_l|+|X_{l+1}|+|C_l|+N_{\mathrm{pub},l})
 +O(q_{\mathrm{sc},l}m_{\max}+HT) \\
 &\qquad +q_{\mathrm{eval},l}\mathsf{VerifyEval}.
\end{aligned}
\label{eq:verifier-complexity}
\end{equation}
The correction and witness terms are linear because the present construction
publishes those values. This linear work is delegated to the worker rather than
eliminated; sublinear checking would add commitments and proofs for these
values. Without batching,
\eqref{eq:fw-block} has eight committed weight tensors per block. The GKR
portion can be streamed with $O(m_{\max})$ field elements per active sum-check.
The verifier stores model commitments, public boundaries, and either the
transcript or its running hash, while the worker reads the disclosed public
witness.

For the HND-32M real block,
$N_{\mathrm{pub}}=259{,}728$ signed integers (255,488 in the numeric trace),
$q_{\mathrm{sc}}=22$, and only six terminal claims require committed-weight
openings. The retained verification path is therefore
sub-matrix-multiplication, but the delegated worker path remains linear in the
public witness. Because the query and signed response are block-local,
different blocks can be assigned to separate workers. Such scheduling does not
reduce aggregate work. Composing an $L$-block chain requires $O(L)$ signature
checks and adjacent-boundary comparisons in addition to the block-local proof
checks.

\subsection{Verification Guarantees}\label{sec:soundness}

For honest execution, each matrix and pointwise reduction is an identity of
multilinear extensions. The sum-check messages therefore satisfy their round
equations, folding preserves the claimed linear combination, registered
evaluations pass $\mathsf{VerifyEval}$, and each block reduction reaches its
public input boundary. The implementation rejects a zero terminal folding
coefficient. If fold $f$ uses $m_f$ Boolean variables, the resulting honest
failure probability is at most $\sum_f(m_f+1)/|\Fp|$ by Schwartz--Zippel and a
union bound.

Algebraic integrity requires the model commitments, request identifier, public
boundaries, and corrections to be fixed before the first challenge. Let
$m_{\mathrm{in}}$ and $m_{\mathrm{out}}$ be the numbers of Boolean variables
in the boundary MLEs. Let $\mathcal S$ be the set of sum-check instances; let
$n_s$ and $\delta_{s,j}$ denote the number of rounds and individual degree in
round $j$; and let $q_{\mathrm{fold}}^{\mathrm{tot}}$ and
$q_{\mathrm{eval}}^{\mathrm{tot}}$ denote the total numbers of folds and
accepted weight evaluations across the accepted block chain. Writing $H_{W,C}$
for the composed registered relation in \eqref{eq:hnd-verification-target}, a
prover that controls its opening helper satisfies
\begin{equation}
\begin{aligned}
 \Pr[\mathsf{Accept}\wedge X_L\ne H_{W,C}(X_0)] \le{}&
 \frac{m_{\mathrm{in}}+m_{\mathrm{out}}+q_{\mathrm{fold}}^{\mathrm{tot}}}{|\Fp|} \\
 &+\frac{\sum_{s\in\mathcal S}\sum_{j=1}^{n_s}\delta_{s,j}}{|\Fp|}
 +q_{\mathrm{eval}}^{\mathrm{tot}}\epsilon_{\mathrm{eval}}.
\end{aligned}
\label{eq:soundness-bound}
\end{equation}
If the public output is not the result of the registered relation, backward
reduction in at least one block reaches a false polynomial claim. A fresh random challenge
accepts a nonzero discrepancy only with its Schwartz--Zippel/sum-check bound.
Random linear folding cancels a false claim with probability at most
$1/|\Fp|$ per completed fold. A zero terminal coefficient is rejected and is
not an accepting event. Otherwise
the false claim reaches a registered tensor, where acceptance implies a failed
evaluation-soundness event. A union bound gives \eqref{eq:soundness-bound}.
This cryptographic bound holds when $V$ executes the prescribed verifier
algorithm; it does not require the prover-side opening helper to be independent.

The disclosed fixed-point relation combines that algebraic argument with a
deterministic public scan. The registrar must derive the pinned model identifier
from the same signed tables used for the eight commitments. If the assigned
worker validates every bound and rescale witness in
\eqref{eq:fixed-schedule}, the 22 matrix reductions imply the signed
fixed-point block relation subject to their sum-check and six-opening error
bound. Both the bounded integer dot product and $sQ+R$ remain below $p/2$, so
field equality lifts to equality of their unique signed representatives. The
scan separately enforces rounding, pointwise products, the mask, and residual
additions. The retained verifier accepts these checks only through the
worker's authenticated response. They impose no additional cryptographic
constraints on the prover, while their linear work remains outside the
verifier process.

For delegated public checks, the retained verifier first validates the
sum-check transcript and derives the terminal query from that transcript. The
assigned worker evaluates the public data independently and signs its response.
If the prover supplies an invalid transcript, verification stops before the
terminal query is issued. If the worker returns an inconsistent value, the
comparison fails. Thus, with an honest retained verifier and at least one honest
party among the prover and assigned worker, an invalid delegated public check is
rejected. Joint control of both parties is outside this non-collusion model; the
registered-weight claims remain protected separately by their ML--KZG openings.

Direct operators can use the same query and response binding when they are
included in the verified statement. Their outputs must refer to the same model,
request, and backbone boundaries. The standalone experiment reported below
evaluates the signed public-computation path; end-to-end direct operators are
kept separate from that result.

The implementation uses the approximately 255-bit BLS12-381 scalar field. Its
Fiat--Shamir transcript expands each challenge to 64 hash bytes before field
reduction and incrementally absorbs the complete statement and preceding
messages. Equation~\eqref{eq:soundness-bound} describes the algebraic error;
the non-interactive implementation additionally relies on the random-oracle
model and the computational binding of the experimental ML--KZG backend. A
production deployment would require an externally authenticated setup ceremony
and an audited commitment implementation; the benchmark setup generated in
this work is not such a ceremony.

\subsection{Communication and Prover Cost}

Supplementary Table~S1 in File~S1 enumerates the per-block verification budget
for \eqref{eq:fw-block}. Model commitment is a one-time registration cost,
whereas evaluation-proof generation occurs after transcript-dependent query
points are known and is therefore online unless the backend supports applicable
preprocessing.

The round count is determined by the contracted or pointwise index. In the
unbatched accounting of Supplementary Table~S1, the operator relations
contribute $q_{\mathrm{base}}=9+4H$ sum-checks and
$q_{\mathrm{eval}}^{\mathrm{blk}}=8$ weight evaluations per block. If the
backward claim graph requires $q_{\mathrm{fold}}^{\mathrm{blk}}$ folding
sum-checks (six in the current implementation when $H>1$), then
$q_{\mathrm{sc}}^{\mathrm{blk}}=q_{\mathrm{base}}+q_{\mathrm{fold}}^{\mathrm{blk}}$.
Thus the
verifier's per-block online work is
\begin{equation}
    O(q_{\mathrm{sc}}^{\mathrm{blk}}m_{\max})
    +O(q_{\mathrm{eval}}^{\mathrm{blk}}\cdot\mathsf{VerifyEval}),
\end{equation}
plus the two boundary MLEs. A future batched-opening construction could reduce
the number of commitment proofs, but no such reduction is claimed or measured
in this paper.

Communication consists of the GKR transcript, the public output, and
backend-specific evaluation proofs. If one field element uses $b_F$ bytes, the
interactive GKR messages occupy
\begin{equation}
    b_F\sum_{s\in\mathcal{S}} n_s(\Delta_s+1)
    +O((|X_L|+q_{\mathrm{fold}}^{\mathrm{tot}})b_F)
\end{equation}
bytes before serialization metadata. The prototype adds one 32-byte value and
one compressed 48-byte G1 quotient commitment per ML--KZG opening variable.
Section~\ref{sec:evaluation} reports an estimated in-memory payload for the
complete synthetic-relation proof, whose versioned canonical framing is not
yet implemented. The real fixed-point path separately uses and parses the
versioned canonical encoding reported in Table~\ref{tab:real-proof}.

The reference prover materializes evaluation tables and invokes one product
sum-check per operator or fold. It is intended to validate the complete claim
graph, not to be an optimized time-optimal GKR implementation. The reported
prover time therefore includes the current table folding and online ML--KZG
opening cost, while setup and model commitment are reported separately.

The interactive protocol has a sequential challenge dependency. The prototype
applies Fiat--Shamir in the random-oracle model~\cite{fiatShamir86} with an
incremental SHA-256 state; this removes verifier messages but not the prover's
internal message--challenge order or transcript size. Each challenge binds the
registered model identifier, public statement, and all preceding messages.

\section{Experimental Analysis}
\label{sec:evaluation}

The experiments separate six questions: whether the implementation realizes
the registered block relation, whether chained fixed-point execution remains
close to the checkpoint path, whether proofs are accepted and tampering is
rejected, whether block proof checks can execute concurrently, whether
the construction reaches the HND-124M block dimensions, and what costs remain
at the prover, retained verifier, computation worker, and result-composition
layers. This separation
prevents numerical agreement, direct recomputation, and protocol concurrency
from being reported as cryptographic coverage of one another.

\subsection{Checkpoint and Protocol Setup}

Strict loading of the backbone and correction artifacts gives the HND-32M
configuration $d=512$, $H=8$, $d_h=64$, $d_{\mathrm{ff}}=2048$, and $L=8$.
We fixed validation indices 0,
1, and 2 before inspecting the chained errors. For each prefix, the original
correction networks produce the per-layer Q/K and RMS correction tables. Their
non-polynomial execution remains outside the proof; the frozen tables are
public statement data.

The source checkpoint, attention-correction network, normalization-correction
network, tokenizer, and validation cache have SHA-256 prefixes
\texttt{3b38ef54}, \texttt{37ae6917}, \texttt{983c3c84},
\texttt{d98595c6}, and \texttt{3faa2b24}, respectively. Python 3.12.12,
PyTorch 2.9.1, and CUDA 12.8 generate the checkpoint-derived exports on RTX
3090 and RTX 5060 Ti GPUs. An independent Rust implementation parses each
binary, re-executes the integer schedule with checked i128 accumulators, and
compares all 38 field-trace digests. The proof implementation uses a
single-threaded Rust 1.97.0 release build on an Intel Xeon Platinum 8173M.

Complete-split likelihood diagnostics for the 32M and larger checkpoints are
reported in the supporting information. They are kept outside the main proof
evaluation because the untouched paths contain non-finite outputs and the
finite values require a disclosed safety-retry rule. Neither perplexity nor the
client-side correction computation is asserted to be proved here.

\subsection{Chained Fixed-Point Agreement}

The chained exporter first quantizes the embedded input and then applies the
source-order schedule in \eqref{eq:fixed-schedule}. For block $l>0$, its input
is the exact signed-integer output of block $l-1$, rather than a newly rounded
FP32 hidden state. Each export records both hashes, and all 21 inter-block links
across the three prefixes match byte for byte. This construction measures error
accumulation while preserving an exact relation between adjacent proved block
instances.

Table~\ref{tab:real-fixedpoint} reports the maximum error over the three
prefixes for every block. The local column compares one fixed-point block with
the same block evaluated on that quantized input. The chained column compares
the accumulated fixed-point output with the original FP32 checkpoint hidden
state at the same depth. Across all 24 exports, the explicit float64 relation
matches the PyTorch block to maximum absolute error at most
$4.63\times10^{-8}$; the independent Rust replay obtains 38/38 digest agreement
for every export. No accumulator exceeds 46 signed-magnitude bits. The largest
chained relative error is $1.36\times10^{-4}$ at block~7.

\begin{table}[htbp]
\centering
\caption{Chained HND-32M fixed-point audit at $T=8$ and Q20. Errors are maxima
over validation indices 0, 1, and 2. Every one of the 24 exports has 38/38
Python--Rust digest agreement and an exact link to the preceding block.}
\label{tab:real-fixedpoint}
\begin{tabular*}{\textwidth}{@{\extracolsep{\fill}}rrrr}
\toprule
Block & Local relative $\ell_2$ & Chained relative $\ell_2$ & Max. accumulator bits \\
\midrule
0 & $6.0708\times10^{-5}$ & $6.0705\times10^{-5}$ & 46 \\
1 & $5.3898\times10^{-5}$ & $7.7345\times10^{-5}$ & 46 \\
2 & $5.1653\times10^{-5}$ & $9.1068\times10^{-5}$ & 46 \\
3 & $4.6684\times10^{-5}$ & $1.0244\times10^{-4}$ & 45 \\
4 & $4.1582\times10^{-5}$ & $1.1379\times10^{-4}$ & 45 \\
5 & $3.4996\times10^{-5}$ & $1.1313\times10^{-4}$ & 45 \\
6 & $3.0575\times10^{-5}$ & $1.2149\times10^{-4}$ & 46 \\
7 & $2.8842\times10^{-5}$ & $1.3635\times10^{-4}$ & 46 \\
\bottomrule
\end{tabular*}
\end{table}

The supporting information reports the precision and prefix-length ablations
used to select Q20. On block~0, Q20 reduces output relative error by about a
factor of 16 compared with Q16 while keeping the maximum accumulator width at
46 bits; Q24 reduces the error further but raises that width to 54 bits. At
fixed Q20, increasing $T$ from 8 to 32 changes the local relative error only
from $5.87\times10^{-5}$ to $6.07\times10^{-5}$. Accepted block-0 proofs at
$T=16$ and $T=32$ are reported in the supporting information. The all-block
proof chain below uses $T=8$.

\subsection{Real-Checkpoint Public-Witness Proof}

For validation index~0, we register and prove each of the eight chained block
instances in Table~\ref{tab:real-fixedpoint}. All blocks use the same public
SRS, but each registration commits to its own six weight matrices and two
$\gamma$ vectors and produces a distinct model identifier. All eight proofs are
accepted, have distinct SHA-256 digests, and retain 38/38 Python--Rust trace
agreement. Equality between a public output tensor and the next public input
tensor links adjacent statements. Thus the experiment proves the chained
polynomial backbone for this prefix under eight block-specific registrations;
it does not prove the generation of the correction tables or the surrounding
language-model operators.

The retained verifier receives the public SRS, registration, public witness,
proof, and pinned model identifier. It verifies the artifact through registered
commitments rather than dense weights or the fixed-point \texttt{forward}
function. For
each block, the 2,077,872-byte witness contains 259,728 signed integers,
including the numeric trace. The current chain stores a 19,092-byte proof for
each block. The earlier block-0 benchmark retained below uses an 18,780-byte
encoding; both proof paths check six committed-weight and sixteen public-table
matrix relations. The two $\gamma$ vectors are
registrar-authenticated public data bound into the model identifier. The legacy
proof digest fixes the input and correction tables but intentionally does not
pin the prover's claimed output or contain a nonce. In the authenticated worker
path, the verifier derives a separate canonical query after checking the
transcript. That query binds the model identifier, request digest, block,
operation, transcript digest, numeric policy, freshness nonce, and evaluation
points. The worker signs the complete canonical response. This mechanism
authenticates the delegated public evaluation; proof-level request binding is a
separate requirement.

To obtain controlled performance statistics, we retain the original block-0
experiment on three independently selected prefixes. These measurements use
one registration and one shared SRS and repeat proving and verification five
times per prefix; they are not multiplied by eight to estimate end-to-end
latency.

\begin{table}[htbp]
\centering
\caption{HND-32M block-0 performance on three prefixes. Proving and
verification are means of five consecutive runs per prefix; parse/hash is one
run. Times are seconds.}
\label{tab:real-proof}
\resizebox{\linewidth}{!}{%
\begin{tabular}{rrrrrrrr}
\toprule
Validation index & Output relative $\ell_2$ & Prove & Parse/hash & Verify incl. scan & Online total & Proof bytes & Accepted \\
\midrule
0 & $5.8705\times10^{-5}$ & 40.948 & 0.0278 & 0.1834 & 0.2112 & 18,780 & yes \\
1 & $5.9279\times10^{-5}$ & 40.897 & 0.0232 & 0.1811 & 0.2043 & 18,780 & yes \\
2 & $6.0708\times10^{-5}$ & 40.999 & 0.0250 & 0.1815 & 0.2065 & 18,780 & yes \\
\bottomrule
\end{tabular}%
}
\end{table}

The one-time fresh SRS generation, block-0 registration, and parameter
serialization take 691.804, 63.511, and 34.130~s, respectively. Canonically
loading and validating the 125,883,136-byte SRS and 8,796-byte registration
takes 276.915~s and can be cached across requests. An isolated standalone
verifier, invoked with no exporter or dense-weight input, accepts index~0 after
this cold load and then requires 0.2063~s per instance.

Across the 15 deployment-path repetitions, proof generation averages
40.948~s and verification including the public numeric scan averages 0.1820~s.
Adding the 0.0253-s mean parse/hash cost gives a 0.2073-s online total. Direct
fixed-point replay with resident dense weights averages 0.09072~s under the
same repeated experiment. The unoptimized verifier is therefore
$2.29\times$ slower than replay and does not demonstrate a latency advantage.
Its present benefit is an integrity check against pinned registrations without
dense-weight storage or matrix replay. The supporting information records the
full timing distributions, peak memory, an earlier unexplained 81.045-s proof
run, and all raw samples; no speedup is inferred from the driver disparity.

\subsection{Parallel Proof Scheduling and HND-124M Extension}

We first compare proof verification and direct fixed-point replay under the same
parallel scheduler. The driver loads the same eight HND-32M blocks for both
paths and checks every artifact before timing. After three warm-up runs, it
records 15 runs for each thread count and alternates which path is measured
first. The 280.144-s artifact load and preflight check are excluded.
Table~\ref{tab:fair-parallel} reports the medians. From one to eight threads,
proof verification improves by $7.45\times$, while replay improves by
$6.27\times$. Verification remains slower than replay; at eight threads, the
ratio is $1.69\times$.

\begin{table}[htbp]
\centering
\caption{Fair same-host HND-32M parallel comparison. Each entry is the median
of 15 post-warm-up runs over the same eight blocks; times are seconds.}
\label{tab:fair-parallel}
\begin{tabular*}{\textwidth}{@{\extracolsep{\fill}}rrrrr}
\toprule
Threads & Proof verify & Direct replay & Proof speedup & Proof/replay \\
\midrule
1 & 1.555248 & 0.773003 & 1.000 & 2.012 \\
2 & 0.829382 & 0.407903 & 1.875 & 2.033 \\
4 & 0.433252 & 0.229410 & 3.590 & 1.889 \\
8 & 0.208697 & 0.123241 & 7.452 & 1.693 \\
\bottomrule
\end{tabular*}
\end{table}

The authenticated process test assigns one Ed25519 identity to the computation
worker. The retained verifier accepts the honest signed responses and rejects a
wrong key, a forged signature, a duplicated response, and replay under a
different query binding. A canonical signed response occupies 397 bytes for one
field value and 429 bytes for two values. These measurements describe the
single-worker protocol.

The direct embedding, correction-generation, and final-output paths are not
included in this standalone authenticated worker result. Their evaluation is
reported separately from the polynomial-backbone and public-witness checks.

We next evaluate a checkpoint-derived HND-124M instance with $d=768$, $H=12$,
$d_{\mathrm{ff}}=3072$, $L=12$, $T=8$, and Q20 semantics. Three clean
repetitions each accept all 12 block proofs and confirm all 11 adjacent
boundaries exactly. The median shared-SRS load, registration, proof generation,
and online chain verification times are 1110.406, 632.007, 1093.443, and
2.812~s, respectively; the median end-to-end time is 2841.731~s. Each block
uses a 23,932-byte proof and a 3,116,720-byte public witness, giving totals of
287,184 and 37,400,640 bytes. Median peak resident memory is 3,567,272~KiB.
This experiment uses an in-process worker adapter and measures feasibility, not
cross-host latency.

Finally, we measure the HND-124M operators assigned to direct validation on an
RTX 3090. After 10 warm-up runs, 50 repetitions give a summed component mean of
0.015848~s for embedding lookup, correction evaluation, final normalization,
the all-token language-model head, and greedy selection at $T=8$. The 3.235-s
model load is excluded. The checkpoint path and the Q20 final boundary select
the same next token. This benchmark measures deterministic recomputation, not
cryptographic proof coverage, and remains separate from the authenticated
worker experiment.

\subsection{Synthetic Scaling}

A separate experiment executes the complete multi-layer relation in
\eqref{eq:fw-block} on fixed-seed BLS12-381 field tensors and correction tables.
Each run generates an experimental SRS, registers all layer tensors, and pins
the resulting model identifier before proving. After one warm-up, proving and
verification are repeated 20 times. Table~\ref{tab:protocol-benchmark} reports
the means. Because this path predates the versioned fixed-point encoding, its
payload is the sum of compressed proof fields rather than a canonical wire
size. The benchmark validates the claim graph and its scaling trend; it is not
a substitute for the checkpoint experiment or an authenticated setup ceremony.

\begin{table}[htbp]
\centering
\caption{Commitment-composed protocol benchmark. Times are seconds; prove and
verify are means of 20 accepted post-warm-up runs. Setup and commitment are offline.}
\label{tab:protocol-benchmark}
\resizebox{\linewidth}{!}{%
\begin{tabular}{rrrrrrrrr@{\hspace{1.5em}}r@{\hspace{1.5em}}r}
\toprule
$T$ & $d$ & $H$ & $d_{\mathrm{ff}}$ & $L$ & Trace & Setup & Commit & Prove & Verify & Est. proof (KiB) \\
\midrule
32  & 64  & 8 & 128 & 2 & 0.091 & 6.789 & 1.012 & 2.600 & 0.279 & 99.5 \\
64  & 64  & 8 & 128 & 2 & 0.205 & 6.927 & 1.028 & 2.900 & 0.288 & 112.0 \\
128 & 64  & 8 & 128 & 2 & 0.499 & 6.790 & 1.020 & 3.658 & 0.288 & 124.5 \\
32  & 128 & 8 & 256 & 4 & 0.659 & 27.713 & 8.278 & 16.975 & 0.610 & 210.9 \\
\bottomrule
\end{tabular}%
}
\end{table}

Increasing $T$ from 32 to 128 raises prover time from 2.600 to 3.658~s and the
estimated payload from 99.5 to 124.5~KiB; verifier time changes from 0.279 to
0.288~s. The fourth row is not a pure scaling point: it jointly doubles $d$ and
$d_{\mathrm{ff}}$ and doubles $L$, raising prove time to 16.975~s. These
measurements validate execution of the claim graph and expose the prototype's
cost profile; they do not predict checkpoint-scale latency.

\subsection{Robustness and Limitations}

The frozen artifact release \texttt{gkr-hnd-test-suite-v1} contains 142 Rust
tests: 66 unit tests and 76 integration tests across 13 integration targets.
The unit tests exercise the arithmetic relation and the bindings used by the
verifier. The integration tests extend this coverage to malformed artifacts
and to protocol paths that cross process or network boundaries. The archived
release output and a machine-readable map identify the claim checked by every
test. These tests detect regressions; they are not a statistical estimate of
cryptographic soundness and do not replace audit or fuzzing.

The checkpoint experiment proves one HND-32M $T=8$ polynomial-backbone chain as
eight block-specific statements. The other two prefixes provide all-block
numerical evidence only. The $T=16$ and $T=32$ experiments have accepted
block-0 proofs, but no all-block proof chain. For HND-124M, all 12 block-specific
proofs are accepted and all 11 adjacent links are audited. Both model
sizes still use public tensor equality between adjacent blocks rather than one
aggregated or recursive multi-block proof.

Embeddings, correction generation, final RMSNorm, the language-model head, and
output selection remain outside the GKR relation. Their deterministic costs are
measured separately, but the current authenticated worker path does not bind
their results to the block proof. The authenticated experiment uses one local
worker process with a pinned Ed25519 key. Cross-host transport, remote
attestation, straggler control, and a network adversary remain future evaluation
targets. The public witness exposes the complete numeric trace, and the worker
scans it linearly. Parallel block assignment can change wall-clock scheduling
without reducing aggregate work, but is not part of the single-worker
authentication result.

The evaluation ML--KZG backend also has a large SRS. The serialized HND-32M
parameters occupy 125,883,136 bytes, and setup reaches about 0.80~GiB peak
resident memory. The HND-124M serialized SRS occupies 503,420,128 bytes. Across
the three clean HND-124M repetitions, median peak
resident memory is 3,567,272~KiB. The experimental setup is generated in-process rather
than by a certified ceremony, and online openings depend on preceding
Fiat--Shamir challenges. Finally, comparison with zkGPT, zkLLM, ZKML, or
Mystique would require a common relation, public-witness policy, security level,
hardware, and wire format. We claim neither one end-to-end cryptographic proof
of language-model inference nor a speedup over those systems. The implemented
result is a same-host authenticated public-evaluation decision combined with
separately validated block proofs. Production networking and end-to-end
composition remain future work.

\section{Conclusion}
\label{sec:conclusion}

This paper presented GKR-HND as a verification protocol for registered HND
Transformer backbones. A retained verifier checks the GKR transcript and
registered-weight openings, while an assigned computation worker evaluates the
remaining public queries. Assuming an honest retained verifier and
prover--worker non-collusion, the verifier
accepts only a signed worker response that matches the proof-derived query.
Each proof remains block-local, preserving compatibility with later recursive
or PoH-style composition.

The prototype links three checkpoint-derived numerical chains across all eight
HND-32M blocks and obtains exact Python--Rust digest agreement. For one chain,
eight block-specific registered proofs are accepted and adjacent public states
match exactly; the maximum accumulated relative error is
$1.36\times10^{-4}$. A second checkpoint-derived experiment accepts all 12
HND-124M block proofs and all 11 adjacent links. The standalone worker path
authenticates a request-bound public evaluation and rejects a wrong key, forged
signature, duplicate response, or replayed query. This path currently uses a
single same-host worker and remains separate from the checkpoint chain
experiment.
Public-witness linearity, the large experimental SRS, certified setup,
cross-host deployment, direct-operator binding, and recursive aggregation
remain open. The artifact establishes block-proof feasibility at the reported
model scales and authenticates delegated public evaluation in a separate
same-host process.

\section*{Acknowledgments}
This work was supported in part by the National Key Research
and Development Program of China under Grant 2025YFE0216300. The authors
declare no competing interests. During manuscript revision, the authors used
OpenAI Codex to assist with structure and language; the authors reviewed and
edited the resulting text and take full responsibility for the content.

\section*{Supporting Information and Artifact Scope}
The supporting information accompanying this preprint reports the precision
and prefix-length ablations, the checkpoint-stability diagnostic, additional
proof-chain results, and the artifact trust boundary. Complete pretrained
checkpoints, correction networks, tokenizers, and validation caches are not
redistributed. The replay archives and large public inputs described in the
artifact accounting are not included in this arXiv source bundle; their byte
sizes and SHA-256 identities are recorded in the supporting information.



\begingroup
\begin{thebibliography}{99}

\bibitem{zheng2025review}
Y.~Zheng, Y.~Chen, B.~Qian, X.~Shi, Y.~Shu, and J.~Chen,
``A review on edge large language models: Design, execution, and applications,''
\textit{ACM Comput. Surv.}, vol.~57, no.~8, pp.~1--35, 2025,
doi: 10.1145/3719664.

\bibitem{gong2023edge}
T.~Gong, L.~Zhu, F.~R. Yu, and T.~Tang,
``Edge intelligence in intelligent transportation systems: A survey,''
\textit{IEEE Trans. Intell. Transp. Syst.}, vol.~24, no.~9,
pp.~8919--8944, Sep. 2023, doi: 10.1109/TITS.2023.3275741.

\bibitem{he2024llms}
Y.~He, J.~Fang, F.~R.~Yu, and V.~C.~Leung,
``Large language models (LLMs) inference offloading and resource allocation in
cloud-edge computing: An active inference approach,''
\textit{IEEE Trans. Mobile Comput.}, vol.~23, no.~12, pp.~11253--11264, 2024,
doi: 10.1109/TMC.2024.3415661.

\bibitem{safetynets2017}
Z.~Ghodsi, T.~Gu, and S.~Garg,
``SafetyNets: Verifiable execution of deep neural networks on an untrusted cloud,''
in \textit{Proc. NeurIPS}, 2017.

\bibitem{zhao2021veriml}
L.~Zhao, Q.~Wang, C.~Wang, Q.~Li, C.~Shen, and B.~Feng,
``VeriML: Enabling integrity assurances and fair payments for machine learning as a service,''
\textit{IEEE Trans. Parallel Distrib. Syst.}, vol.~32, no.~10,
pp.~2524--2540, Oct. 2021, doi: 10.1109/TPDS.2021.3068195.

\bibitem{ggpr13}
R.~Gennaro, C.~Gentry, B.~Parno, and M.~Raykova,
``Quadratic span programs and succinct NIZKs without PCPs,''
in \textit{Proc. EUROCRYPT}, 2013.

\bibitem{benSasson2018stark}
E.~Ben-Sasson, I.~Bentov, Y.~Horesh, and M.~Riabzev,
``Scalable, transparent, and post-quantum secure computational integrity,''
\textit{IACR Cryptol. ePrint Arch.}, Rep.~2018/046, 2018.

\bibitem{spartan2020}
S.~Setty,
``Spartan: Efficient and general-purpose zkSNARKs without trusted setup,''
in \textit{Proc. CRYPTO}, 2020.

\bibitem{hnd}
X.~Liang, Y.~Lv, J.~Li, R.~Qin, Y.~Tian, and F.-Y.~Wang,
``Homomorphic-Nonhomomorphic Decomposition: A Verification-Friendly
Transformer Architecture,''
SSRN, 2026, doi: 10.2139/ssrn.6946183.

\bibitem{gkr}
S.~Goldwasser, Y.~T.~Kalai, and G.~N.~Rothblum,
``Delegating computation: interactive proofs for muggles,''
in \textit{Proc. STOC}, 2008.

\bibitem{thaler2013}
J.~Thaler,
``Time-optimal interactive proofs for circuit evaluation,''
in \textit{Proc. CRYPTO}, 2013.

\bibitem{babai1988arthur}
L.~Babai and S.~Moran,
``Arthur--Merlin games: a randomized proof system, and a hierarchy of
complexity classes,''
\textit{J. Comput. Syst. Sci.}, vol.~36, no.~2, pp.~254--276, 1988.

\bibitem{shamir1992ip}
A.~Shamir,
``IP = PSPACE,''
\textit{J. ACM}, vol.~39, no.~4, pp.~869--877, 1992.

\bibitem{thaler2022}
J.~Thaler,
\textit{Proofs, Arguments, and Zero-Knowledge},
\textit{Found. Trends Privacy Secur.}, vol.~4, no.~2--4, pp.~117--660, 2022,
doi: 10.1561/3300000030.

\bibitem{lfkn}
C.~Lund, L.~Fortnow, H.~Karloff, and N.~Nisan,
``Algebraic methods for interactive proof systems,''
\textit{J. ACM}, vol.~39, no.~4, pp.~859--868, 1992.

\bibitem{cormode2011}
G.~Cormode, J.~Thaler, and K.~Yi,
``Verifying computations with streaming interactive proofs,''
\textit{Proc. VLDB Endowment}, vol.~5, no.~1, pp.~25--36, 2011,
doi: 10.14778/2047485.2047488.

\bibitem{flow2019verification}
A.~Hentschel, D.~Shirley, L.~Lafrance, and M.~Zamski,
``Flow: Separating consensus and compute---execution verification,''
\textit{arXiv preprint arXiv:1909.05832}, 2019.

\bibitem{pinocchio13}
B.~Parno, J.~Howell, C.~Gentry, and M.~Raykova,
``Pinocchio: Nearly practical verifiable computation,''
in \textit{Proc. IEEE Symp. Secur. Privacy}, 2013.

\bibitem{aurora2019}
E.~Ben-Sasson, A.~Chiesa, M.~Riabzev, N.~Spooner, M.~Virza, and N.~P.~Ward,
``Aurora: Transparent succinct arguments for R1CS,''
in \textit{Proc. EUROCRYPT}, 2019.

\bibitem{deepfri2020}
E.~Ben-Sasson, L.~Goldberg, S.~Kopparty, and S.~Saraf,
``DEEP-FRI: Sampling outside the box improves soundness,''
in \textit{Proc. ITCS}, 2020.

\bibitem{groth16}
J.~Groth,
``On the size of pairing-based non-interactive arguments,''
in \textit{Proc. EUROCRYPT}, 2016.

\bibitem{kzg10}
A.~Kate, G.~M.~Zaverucha, and I.~Goldberg,
``Constant-size commitments to polynomials and their applications,''
in \textit{Proc. ASIACRYPT}, 2010.

\bibitem{bootle2016}
J.~Bootle, A.~Cerulli, P.~Chaidos, J.~Groth, and C.~Petit,
``Efficient zero-knowledge arguments for arithmetic circuits in the discrete log setting,''
in \textit{Proc. EUROCRYPT}, 2016.

\bibitem{marlin2020}
A.~Chiesa, Y.~Hu, M.~Maller, P.~Mishra, N.~Vesely, and N.~P.~Ward,
``Marlin: Preprocessing zkSNARKs with universal and updatable SRS,''
in \textit{Proc. EUROCRYPT}, 2020.

\bibitem{dory2021}
J.~Lee,
``Dory: Efficient, transparent arguments for generalised inner products and polynomial commitments,''
in \textit{Proc. TCC}, 2021.

\bibitem{fri2018}
E.~Ben-Sasson, I.~Bentov, Y.~Horesh, and M.~Riabzev,
``Fast Reed--Solomon interactive oracle proofs of proximity,''
in \textit{Proc. ICALP}, 2018.

\bibitem{bulletproofs2018}
B.~B{\"u}nz, J.~Bootle, D.~Boneh, A.~Poelstra, P.~Wuille, and G.~Maxwell,
``Bulletproofs: Short proofs for confidential transactions and more,''
in \textit{Proc. IEEE Symp. Secur. Privacy}, 2018.

\bibitem{mystique2021}
C.~Weng, K.~Yang, X.~Xie, J.~Katz, and X.~Wang,
``Mystique: Efficient conversions for zero-knowledge proofs with applications
to machine learning,'' in \textit{Proc. 30th USENIX Security Symp.},
pp.~501--518, 2021.

\bibitem{zkml2024}
B.-J.~Chen, S.~Waiwitlikhit, I.~Stoica, and D.~Kang,
``ZKML: An optimizing system for ML inference in zero-knowledge proofs,''
in \textit{Proc. 19th ACM European Conf. Computer Systems (EuroSys)},
pp.~560--574, 2024, doi: 10.1145/3627703.3650088.

\bibitem{hao2024}
M.~Hao, H.~Chen, H.~Li, C.~Weng, Y.~Zhang, H.~Yang, and T.~Zhang,
``Scalable zero-knowledge proofs for non-linear functions in machine
learning,'' in \textit{Proc. 33rd USENIX Security Symp.}, pp.~3819--3836,
2024.

\bibitem{zkllm2024}
H.~Sun, J.~Li, and H.~Zhang,
``zkLLM: Zero knowledge proofs for large language models,''
in \textit{Proc. ACM CCS}, pp.~4405--4419, 2024,
doi: 10.1145/3658644.3670334.

\bibitem{zkgpt2025}
W.~Qu, Y.~Sun, X.~Liu, T.~Lu, Y.~Guo, K.~Chen, and J.~Zhang,
``zkGPT: An efficient non-interactive zero-knowledge proof framework for LLM
inference,'' in \textit{Proc. 34th USENIX Security Symp.}, pp.~2045--2063,
2025.

\bibitem{ivakhnenko1971polynomial}
A.~G. Ivakhnenko,
``Polynomial theory of complex systems,''
\textit{IEEE Trans. Syst., Man, Cybern.}, vol.~SMC-1, no.~4,
pp.~364--378, Oct. 1971.

\bibitem{hubara2018quantized}
I.~Hubara, M.~Courbariaux, D.~Soudry, R.~El-Yaniv, and Y.~Bengio,
``Quantized neural networks: Training neural networks with low precision weights and activations,''
\textit{J. Mach. Learn. Res.}, vol.~18, no.~187, pp.~1--30, 2018.

\bibitem{jacob2018quantization}
B.~Jacob et al.,
``Quantization and training of neural networks for efficient integer-arithmetic-only inference,''
in \textit{Proc. CVPR}, 2018.

\bibitem{katharopoulos2020linear}
A.~Katharopoulos, A.~Vyas, N.~Pappas, and F.~Fleuret,
``Transformers are RNNs: Fast autoregressive transformers with linear attention,''
in \textit{Proc. ICML}, 2020.

\bibitem{wang2020linformer}
S.~Wang et al.,
``Linformer: Self-attention with linear complexity,''
\textit{arXiv preprint arXiv:2006.04768}, 2020.

\bibitem{kitaev2020reformer}
N.~Kitaev, L.~Kaiser, and A.~Levskaya,
``Reformer: The efficient transformer,''
in \textit{Proc. ICLR}, 2020.

\bibitem{choromanski2021performer}
K.~Choromanski et al.,
``Rethinking attention with performers,''
in \textit{Proc. ICLR}, 2021.

\bibitem{su2024roformer}
J.~Su, M.~Ahmed, Y.~Lu, S.~Pan, W.~Bo, and Y.~Liu,
``RoFormer: Enhanced transformer with rotary position embedding,''
\textit{Neurocomputing}, vol.~568, Art.~127063, 2024,
doi: 10.1016/j.neucom.2023.127063.

\bibitem{fiatShamir86}
A.~Fiat and A.~Shamir,
``How to prove yourself: Practical solutions to identification and signature problems,''
in \textit{Proc. CRYPTO}, 1986.

\end{thebibliography}
\endgroup
\end{document}


\title{Supporting Information for Agree on the Model, Verify the Inference:
GKR Protocols for HND-Based Transformer Inference}
\author{Xiaolong Liang, Juanjuan Li, Rui Qin, and Yisheng Lv\\[0.5em]
\small State Key Laboratory of Multimodal Artificial Intelligence Systems,\\[-0.1em]
\small Institute of Automation, Chinese Academy of Sciences, Beijing 100190, China\\
\small \texttt{yisheng.lv@ia.ac.cn}}
\date{}

\maketitle

\begin{appendix}

\section{Precision and Prefix-Length Ablations}

The main paper uses signed uniform Q20 semantics and a prefix length of
$T=8$ for checkpoint-derived proofs. Table~\ref{tab:s-precision} reports the
ablations used to select that operating point. The exporter performs
nearest-half-away-from-zero rounding and rescales after every source-order
operation. The errors compare the fixed-point output with the explicit
unquantized float64 block relation. The independent Rust replay obtains 38/38
field-digest agreement in every row.

\begin{table}[htbp]
\centering
\caption{HND-32M block-0 precision and prefix-length audit on validation
index~0.}
\label{tab:s-precision}
\begin{tabular*}{\textwidth}{@{\extracolsep{\fill}}rrrrrrr}
\toprule
$T$ & Fraction bits & Relative $\ell_2$ & Max. abs. error & Cosine & Max. bits & Digests \\
\midrule
8  & 16 & $9.6721\times10^{-4}$ & $2.8896\times10^{-5}$ & 0.999999533 & 38 & 38/38 \\
8  & 20 & $5.8705\times10^{-5}$ & $1.5789\times10^{-6}$ & 0.999999998 & 46 & 38/38 \\
8  & 24 & $3.7795\times10^{-6}$ & $1.1302\times10^{-7}$ & 1.000000000 & 54 & 38/38 \\
16 & 20 & $6.0057\times10^{-5}$ & $1.7961\times10^{-6}$ & 0.999999998 & 46 & 38/38 \\
32 & 20 & $6.0703\times10^{-5}$ & $1.8805\times10^{-6}$ & 0.999999998 & 46 & 38/38 \\
\bottomrule
\end{tabular*}
\end{table}

Q20 reduces the output relative error by approximately a factor of 16 compared
with Q16, while all observed accumulators remain within 46 signed-magnitude
bits. Q24 reduces the error further but increases the maximum width to 54 bits.
At fixed Q20, increasing $T$ from 8 to 32 changes the relative error only from
$5.87\times10^{-5}$ to $6.07\times10^{-5}$.

Table~\ref{tab:prefix-proof} records the corresponding block-0 proof artifacts
for validation index~0. All three instances reuse the same public SRS and pass
38/38 cross-language trace checks. Each of the 11 prescribed mutations is
rejected at $T=16$ and $T=32$, as at $T=8$. Doubling the prefix approximately
doubles the public-witness encoding, whereas each additional token-index round
adds only 800 bytes to the canonical proof. These longer-prefix results cover
block~0 only; the all-eight-block proof chain in the main paper uses $T=8$.

\begin{table}[htbp]
\centering
\caption{Accepted HND-32M block-0 Q20 proof artifacts for validation index~0.}
\label{tab:prefix-proof}
\begin{tabular*}{\textwidth}{@{\extracolsep{\fill}}rrrrrr}
\toprule
$T$ & Public witness (bytes) & Proof (bytes) & Digests & Max. bits & Accepted \\
\midrule
8  & 2,077,872 & 19,092 & 38/38 & 46 & yes \\
16 & 4,180,272 & 19,892 & 38/38 & 46 & yes \\
32 & 8,458,800 & 20,692 & 38/38 & 46 & yes \\
\bottomrule
\end{tabular*}
\end{table}

\section{Checkpoint Stability Diagnostic}

This section reports model-quality diagnostics that are deliberately separated
from the proved relation. We use HND-124M-core for the larger checkpoint to
distinguish its Transformer core from the deployed model, which also contains
an untied output projection.

For target tokens $y_1,\ldots,y_N$, the evaluator accumulates FP32 token-level
cross entropy in double precision and reports
\begin{equation}
 \operatorname{PPL}=\exp\!\left(\frac{1}{N}
 \sum_{i=1}^{N}-\log p(y_i\mid y_{<i})\right).
 \label{eq:s-ppl}
\end{equation}
It uses deterministic PyTorch settings, never drops a batch, records exact
token counts, and hashes every model, tokenizer, and validation-cache artifact.
Direct $\alpha=0$ inference produces non-finite logits for some sequences in
both checkpoints, so the untouched full-split PPL is undefined.

The finite diagnostic first executes the original path. Only when one sequence
is non-finite does it retry that sequence after upper-clipping the two
first-block RMS correction tables. Q/K corrections and all later blocks remain
unchanged. The caps are 300 for HND-32M and 50 for HND-124M-core. They were
selected post hoc to restore finite evaluation; they were not preregistered,
and cap sensitivity was not measured. Table~\ref{tab:s-ppl} therefore labels
the results as safety-retry diagnostics rather than untouched checkpoint PPL.

\begin{table}[htbp]
\centering
\small
\caption{Complete validation-split safety-retry perplexity. The retry includes
every token and modifies only sequences whose original logits are non-finite.}
\label{tab:s-ppl}
\begin{tabular*}{\textwidth}{@{\extracolsep{\fill}}lrrlrrr}
\toprule
Model & Sequences & Tokens & Retry cap & Retried & Rate & Retry PPL \\
\midrule
HND-32M & 42,534 & 21,777,408 & first-block RMS $\le300$ & 588 & 1.382\% & 8.0964 \\
HND-124M-core & 52,371 & 53,627,904 & first-block RMS $\le50$ & 13 & 0.0248\% & 19.6547 \\
\bottomrule
\end{tabular*}
\end{table}

The two values are not a controlled size-scaling comparison because the models
use different tokenizers, vocabulary sizes, sequence lengths, and validation
caches. The 124M-core split is also merged across RTX 3090 and RTX 5060 Ti
devices. The merge script rejects overlap, missing indices, artifact-hash
mismatch, and incomplete shards, but deterministic settings do not guarantee
bitwise equality across GPU architectures. No perplexity computation is part
of the GKR-HND proof.

\section{Repeated Timing and Artifact Accounting}

Table~2 of the main paper reports the arithmetic means for three block-0
prefixes. Across the 15 recorded repetitions, proof generation has mean
40.948~s, median 40.796~s, and range 40.662--41.756~s. Verification including
the public numeric scan has mean 0.1820~s, median 0.1817~s, and range
0.1796--0.1882~s. The three parse/hash measurements average 0.0253~s, giving a
0.2073-s mean online total. Direct fixed-point replay with resident dense
weights averages 0.09072~s over 15 runs, with range 0.08861--0.09544~s.

Fresh SRS generation takes 691.804~s; block-0 registration takes 63.511~s; and
serializing the public parameters takes 34.130~s. A full canonical reload of
the 125,883,136-byte SRS and 8,796-byte registration takes 276.915~s and can be
cached. Each proof in the retained block-0 benchmark is 18,780 bytes and each public witness is
2,077,872 bytes. The witness bytes are not included in the proof-size column.

An earlier integrated fresh-setup driver recorded one 81.045-s proof for the
same block-0 statement, whereas the repeated deployment-path driver records
40.948~s and produces the identical proof SHA-256. The drivers ran in different
system states and no controlled experiment isolates the disparity. The main
paper therefore uses the repeated deployment-path measurements and makes no
proving-speedup claim.

For completeness, the current $T=16$ and $T=32$ block-0 proof drivers each ran
once with the same cached SRS. They record prover times of 43.718 and 45.008~s
and verification times of 0.2323 and 0.3272~s, respectively. These
single-run values document the artifacts in Table~\ref{tab:prefix-proof}; they
are not used as a performance comparison or speedup claim.

\section{Chained Proof Audit}

Table~\ref{tab:s-chain} summarizes the machine-checked manifest for the chained
experiment. All eight block-specific proofs use one SRS but distinct registered
model identifiers. The 24 numerical instances cover eight blocks and three
preselected prefixes. Only validation index~0 has a proof for every block; the
other two chains are numerical audits.

\begin{table}[htbp]
\centering
\caption{Chained checkpoint-derived evidence.}
\label{tab:s-chain}
\begin{tabular}{lr}
\toprule
Invariant & Result \\
\midrule
Numerical block instances & 24 \\
Independent Rust imports accepted & 24/24 \\
Field-trace digest agreement & 38/38 per instance \\
Exact adjacent-state links & 21/21 \\
Block-specific proofs accepted for index~0 & 8/8 \\
Distinct model identifiers / proof hashes & 8 / 8 \\
Shared SRS SHA-256 prefix & \texttt{8185c455} \\
Proof bytes per block & 19,092 \\
Witness bytes per block & 2,077,872 \\
Maximum chained relative $\ell_2$ error & $1.3635\times10^{-4}$ \\
Maximum accumulator width & 46 bits \\
\bottomrule
\end{tabular}
\end{table}

For every proved block, the driver separately changes a witness byte, numeric
remainder, proof message, model identifier, weight commitment, input,
correction, output, causal mask, numeric specification, and block-provenance
field. All 88 modified instances are rejected. These deterministic mutations
are regression checks, not a substitute for cryptographic audit or fuzzing.

\section{Parallel Scheduling, Authenticated Worker, and HND-124M}

Table~\ref{tab:s-fair-parallel} gives the controlled scheduler comparison
described in the main paper. The driver uses ordinary host threads to schedule
eight proof-verification or direct-replay tasks. These threads are benchmark
resources rather than the assigned computation worker in the GKR-HND protocol.
All values are medians of 15 post-warm-up repetitions, with three warm-ups and
alternating path order.

\begin{table}[htbp]
\centering
\caption{Fair eight-block HND-32M proof-verification and replay timings.}
\label{tab:s-fair-parallel}
\begin{tabular}{rrrrr}
\toprule
Threads & Proof (s) & Replay (s) & Proof speedup & Proof/replay \\
\midrule
1 & 1.555248 & 0.773003 & 1.000 & 2.012 \\
2 & 0.829382 & 0.407903 & 1.875 & 2.033 \\
4 & 0.433252 & 0.229410 & 3.590 & 1.889 \\
8 & 0.208697 & 0.123241 & 7.452 & 1.693 \\
\bottomrule
\end{tabular}
\end{table}

The authenticated worker experiment uses one local computation-worker process
with a pinned Ed25519 key. The retained verifier derives each query after the
corresponding proof check and accepts only a signed response with the same
model, request, block, operation, transcript, numeric policy, nonce, and
evaluation points. The honest path accepts; a wrong key, forged signature,
duplicate frame, malformed frame, or replayed binding is rejected. Canonical
signed responses occupy 397 bytes for one field value and 429 bytes for two.
This experiment covers the public numeric scan and terminal public-table
evaluations. Direct embedding, correction generation, final normalization,
logits, and output selection are measured separately.

The HND-124M extension contains twelve block-specific registrations, witnesses,
and proofs. All twelve proofs accept, and all eleven adjacent boundaries match.
Table~\ref{tab:s-hnd124-chain} reports the medians from three clean sequential
repetitions. The worker adapter runs in process, so these data establish
all-block feasibility rather than network latency.

\begin{table}[htbp]
\centering
\caption{Complete HND-124M proof-chain audit at $T=8$ and Q20.}
\label{tab:s-hnd124-chain}
\begin{tabular}{lr}
\toprule
Invariant or cost & Result \\
\midrule
Accepted block proofs & 12/12 \\
Exact adjacent boundaries & 11/11 \\
Shared serialized SRS & 503,420,128 bytes \\
Median shared SRS load & 1110.406 s \\
Median registration & 632.007 s \\
Median proof generation & 1093.443 s \\
Median online chain verification & 2.812 s \\
Median end-to-end time & 2841.731 s \\
Proof bytes per block / total & 23,932 / 287,184 \\
Witness bytes per block / total & 3,116,720 / 37,400,640 \\
Median peak resident memory & 3,567,272 KiB \\
\bottomrule
\end{tabular}
\end{table}

The direct HND-124M operators are benchmarked separately on an RTX 3090. After
10 warm-up iterations, 50 measured iterations give a 0.015848-s sum of the
component means for the all-token path at $T=8$; the 3.235-s model load is
excluded. The checkpoint and Q20 final-boundary paths select the same next
token. This measurement covers deterministic recomputation rather than the
cryptographic relation.

\section{Reproduction Boundary}

The companion archives and content-addressed objects described below document
the evaluated artifact snapshot; they are not included in this arXiv source
bundle.

The release separates compact companion archives from large public inputs.
S1 contains the standalone source, tests, documentation, and tracked raw
results. S2 contains the signed registrations for all 20 evaluated blocks and
the two ordered backbone registrations. S3 provides current-format witnesses
and accepted proofs for the three preselected $T=8$ prefixes, whereas S4
provides the complete eight-block and twelve-block chains together with the
$T=16$ and $T=32$ prefix experiments. At release tag
\texttt{gkr-hnd-test-suite-\allowbreak v1}, the standalone source contains 142
Rust tests: 66 unit tests and 76 integration tests across 13 integration
targets. The release also archives the complete Cargo output and a
machine-readable test-to-claim map.

The two public SRS files and the checkpoint-derived fixed-point exports are
stored as content-addressed objects rather than duplicated inside the ZIP
files. Each object index records a retrieval URI, byte count, and SHA-256
digest. Each companion archive contains its own \texttt{PACKAGE\_MANIFEST.json},
which lists every internal file and its digest. The release-level closure
manifest binds all four archive hashes, all external objects, the artifact-
generation commit, and the packaging commit. An automated closure check rejects
a missing object, a dangling package entry, a legacy proof encoding, or a
package containing the registrar's private signing key.

Complete FP32 checkpoints, correction networks, tokenizers, and validation
caches are not redistributed. Their byte sizes and SHA-256 identities are
recorded. Regenerating the chained fixed-point exports or the likelihood
diagnostic therefore requires authorized access to those original artifacts.
The source-checkpoint hash in a registration is a registrar-attested provenance
label; the current proof does not certify the floating-point-to-fixed-point
export procedure. The Q16 and Q24 records are retained only as fixed-point
fidelity diagnostics; they are not inputs to a released cryptographic proof.

Table~\ref{tab:artifact-layers} separates replay of the released public proof
from regeneration of the checkpoint-derived relation. Replay needs no FP32
checkpoint. After building the Rust verifier from S1, a reviewer retrieves the
indexed public SRS and fixed-point export objects and verifies their digests
before using the registrations, witnesses, proofs, and chain decisions in
S2--S4.

\begin{table}[htbp]
\centering
\small
\caption{Released artifact layers and their trust boundary.}
\label{tab:artifact-layers}
\begin{tabular*}{\textwidth}{@{\extracolsep{\fill}}lll}
\toprule
File & Principal contents & FP32 checkpoint required? \\
\midrule
S1 & Source, tests, documentation, tracked raw results & no \\
S2 & 20 signed block registrations, two backbone roots, SRS index & no \\
S3 & Three $T=8$ block-0 witnesses and accepted proofs & no \\
S4 & Eight- and twelve-block chains, $T=16/32$ proofs & no \\
Regeneration & Export and likelihood re-execution & yes \\
\bottomrule
\end{tabular*}
\end{table}

The replay boundary verifies the disclosed fixed-point relation and its
registered weights. It does not independently establish that a private FP32
checkpoint was trained as claimed, that the exporter selected the correct
checkpoint tensors, or that the public correction tables were generated by the
declared correction networks. Those provenance steps remain part of trusted
registration and the recorded artifact audit.

\end{appendix}